\documentclass[10pt,twocolumn,letterpaper]{article}

\usepackage{cvpr}              

\usepackage{graphicx}
\usepackage{amsmath}
\usepackage{amssymb}
\usepackage{booktabs}

\usepackage{multirow}
\usepackage{pifont}
\usepackage{graphicx}
\usepackage{amsmath}
\usepackage{amssymb}
\usepackage{booktabs}
\usepackage{xcolor}
\usepackage{enumitem}
\usepackage{amsfonts}
\usepackage{xparse}
\usepackage{tabu}
\usepackage{subcaption}
\usepackage[accsupp]{axessibility}  

\definecolor{RoseQuartzBg}{HTML}{F7CAC9}
\definecolor{RoseQuartz}{HTML}{F5A798}
\definecolor{Serenity}{HTML}{92A8D1}
\definecolor{OrangeRed}{rgb}{1.0, 0.27, 0.0}
\definecolor{Turquoise}{HTML}{0F4C81}
\definecolor{mint}{rgb}{0.24, 0.71, 0.54}
\definecolor{captioningtask}{HTML}{AE4132}
\definecolor{qatask}{HTML}{0E8088}
\definecolor{temporalprefix}{HTML}{7F00FF}
\definecolor{targettext}{HTML}{3333FF}
\definecolor{prompttext}{HTML}{666666}
\definecolor{videolevel}{HTML}{82B366}
\definecolor{framelevel}{HTML}{6C8EBF}
\definecolor{tokenlevel}{HTML}{D79B00}
\newcommand{\cmark}{\ding{51}}
\newcommand{\xmark}{\ding{55}}

\NewDocumentCommand{\xudong}{ mO{} }{\textcolor{blue}{\textsuperscript{\textit{Xudong}}\textsf{\textbf{\small[#1]}}}}

\NewDocumentCommand{\yulei}{ mO{} }{\textcolor{purple}{\textsuperscript{\textit{Yulei}}\textsf{\textbf{\small[#1]}}}}

\NewDocumentCommand{\andrew}{ mO{} }{\textcolor{cyan}{\textsuperscript{\textit{Andrew}}\textsf{\textbf{\small[#1]}}}}

\usepackage[pagebackref,breaklinks,colorlinks]{hyperref}

\usepackage[capitalize]{cleveref}
\crefname{section}{Sec.}{Secs.}
\Crefname{section}{Section}{Sections}
\Crefname{table}{Table}{Tables}
\crefname{table}{Tab.}{Tabs.}

\def\cvprPaperID{39} 
\def\confName{L3D-IVU}
\def\confYear{2023}

\begin{document}


\title{In Defense of Structural Symbolic Representation \\ for Video Event-Relation Prediction}

\author{%
  Andrew Lu\thanks{Equal contribution.},~
  Xudong Lin\footnotemark[1],~
  Yulei Niu,~
Shih-Fu Chang
\\
  { Columbia University \ \
  }
  \\
  \texttt{\{ayl2148,xudong.lin,yn2338,sc250\}@columbia.edu}
}
\maketitle

\begin{abstract}
 Understanding event relationships in videos requires a model to understand the underlying structures of events (\textit{i.e.} the event type, the associated argument roles, and corresponding entities) and factual knowledge for reasoning. Structural symbolic representation (SSR) based methods directly take event types and associated argument roles/entities as inputs to perform reasoning.
 However, the state-of-the-art video event-relation prediction system shows the necessity of  using continuous feature vectors from input videos; existing methods based solely on SSR inputs fail completely, even when given oracle event types and argument roles. In this paper, we conduct an extensive empirical analysis to answer the following questions: 1) why SSR-based method failed; 2) how to understand the evaluation setting of video event relation prediction properly; 3) how to uncover the potential of SSR-based methods. We first identify suboptimal training settings as causing the failure of previous SSR-based video event prediction models. 
 Then through qualitative and quantitative analysis, we show how evaluation that takes only video as inputs is currently unfeasible, as well as the reliance on oracle event information to obtain an accurate evaluation. Based on these findings, we propose to further contextualize the SSR-based model to an Event-Sequence Model and equip it with more factual knowledge through a simple yet effective way of reformulating external visual commonsense knowledge bases into an event-relation prediction pretraining dataset. The resultant new state-of-the-art model eventually establishes a \textbf{25\%} Macro-accuracy performance boost. 

\end{abstract}

\section{Introduction}
Event understanding has been thoroughly explored in the past decade~\cite{ye2015eventnet,yatskar2016situation,li2020cross,pratt2020grounded,sadhu2021visual,chen2021joint,nguyen-etal-2016-joint, sha2018jointly, liu-etal-2019-neural-cross, liu2020event, yang2022video,ayyubi2022multimodal}. An event is typically represented as a verb (indicating the event type) and several arguments, each of which has a role name and associated entity. Researchers have been devoted to extracting events and labeling the argument roles~\cite{yatskar2016situation,li2020cross,pratt2020grounded,sadhu2021visual,chen2021joint,wang2019hmeae,wang2020neural} in vision, language, and multimodal domains. These event extraction and argument role labeling models build the foundation for higher level understanding of event-relations. Relations between events~\cite{o2016richer,glavavs2014hieve,badgett2016extracting,yao2020weakly,bosselut2019comet,sap2019atomic} have been thoroughly studied in the language domain. A specific relation and event coreference have also been explored in the multimodal setting~\cite{li2020cross,chen2021joint}. However, video event-relation prediction is still a new and challenging task~\cite{sadhu2021visual} requiring both good representations of video events and commonsense knowledge to reason between events.

Structural symbolic representation (SSR) ~\cite{yi2018neural,hudson2019learning,xu2019scene,ji2020action,li2020cross,lin2021vx2text,wanglanguage,lin2022towards} has been widely adopted on various downstream tasks such as visual question answering~\cite{yi2018neural}, image captioning~\cite{xu2019scene}, and action recognition~\cite{ji2020action} for its high interpretability and generalization ability~\cite{yi2018neural, lin2021vx2text, wanglanguage,lin2022towards}. In this work, we refer to SSR as representations consisting of discrete tokens organized with certain structures. 
When applied to event-relation prediction, it is usually formulated as an event type (verb) with associated argument role names and corresponding entities. 
This event-argument SSR has been very effective in predicting event relations in the language domain~\cite{dligach2017neural}. However, SSR-based models for video events are easily biased by the dominant relations in the VidSitu dataset~\cite{sadhu2021visual} and predict the dominant class for all classes, which contradicts the success of existing SSR-based methods on various tasks including event-relation prediction using text.

To answer why SSR-based models fail on video event-relation prediction, we first analyze if there are any possible patterns that the model could have leveraged to avoid being misled (Sec.~\ref{data_analysis}). We discover that even if the model only memorizes the dominant relation for pairs of event types, the model is clearly not supposed to predict only \textit{Enables} (the most frequent relation in the VidSitu dataset~\cite{sadhu2021visual}). Based on this finding, we bootstrap the failed SSR-based model and the state-of-the-art model~\cite{sadhu2021visual} with two simple processes: utilizing a balanced training data/objective and tuning hyper-parameters. Through these optimized training settings, the text-only SSR-based model outperforms the multimodal state-of-the-art model by 20\%. 

The different behaviors motivate us to carefully analyze the evaluation challenges of video event-relation prediction (Sec~\ref{sec:challenge}). We evaluate two state-of-the-art video-language models, HERO~\cite{li2020hero} and ClipBERT~\cite{lei2021less} on the VidSitu dataset~\cite{sadhu2021visual}. By controlling inputs with the help of strong image-text contrastive models~\cite{radford2021learning}, we found that including oracle information is essential due to the presence of multiple events co-occurring simultaneously in the same video. Even with strong pretrained video-language models, the event-type and argument role descriptions are still more important than video feature vectors.

Based on these observations, we propose to contextualize the simple pairwise SSR-based models to an Event-Sequence model in order to leverage context information within sequences of events for more accurate event-relation prediction. Furthermore, we explore leveraging an external visual knowledge base, VisualCOMET~\cite{park2020visualcomet}, to teach the model commonsense knowledge about the evolution process of events. We propose an effective and straightforward strategy for reformulating the annotations of VisualCOMET into event sequences suitable for event-relation prediction. The contextualized Event-Sequence model with pretraining on VisualCOMET outperforms the state-of-the-art model~\cite{sadhu2021visual} by a \textbf{25\%} improvement in Macro-accuracy.

\vspace{3pt}
Our contributions are summarized as follows:
\vspace{-6pt}
\begin{itemize}[leftmargin=*]
    \item We identify why SSR-based models fail on video event-relation prediction and improve Macro-accuracy performance by 20\% through optimizing training settings.
\vspace{-6pt}
    \item We identify the proper settings needed to evaluate video event-relation prediction models through extensive quantitative and qualitative analysis of different model variations on the VidSitu dataset.
\vspace{-6pt}
    \item We propose a contextualized Event-Sequence model, coupled with a pretraining technique on VisualCOMET, to fully utilize the rich contextual information in event sequences and commonsense knowledge from the existing knowledge base. Our model significantly improves event-relation prediction accuracy compared to the state-of-the-art model from $34.2\%\rightarrow59.2\%$.

\end{itemize}

\begin{figure*}[h!]
    \centering
    \includegraphics[width=0.8\textwidth]{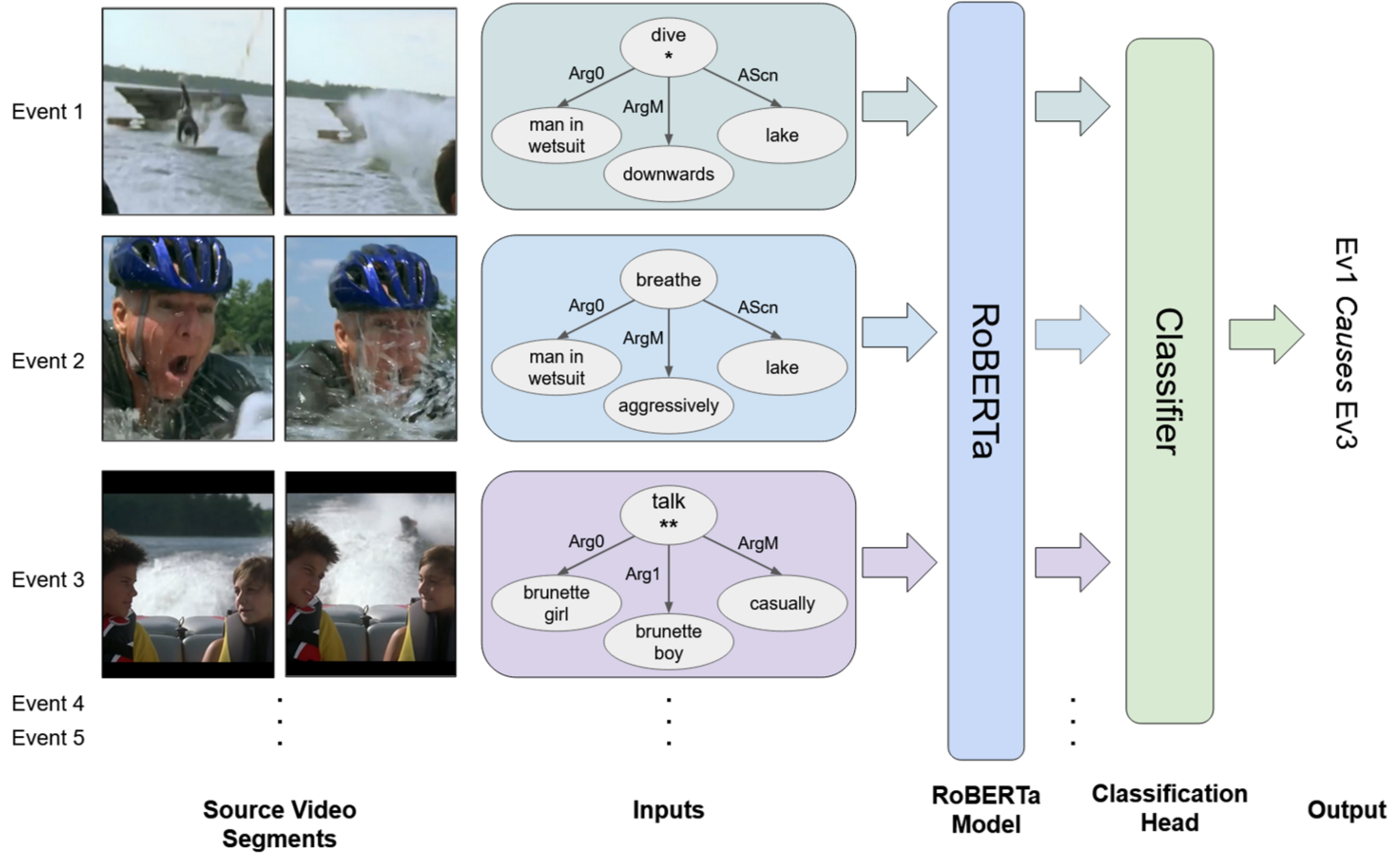}
    \caption{The pipeline of the Event-Sequence model for event-relation prediction. Special characters * and ** denote the target events the model is required to predict relation between. Detailed description is in Sec.~\ref{csm}}
    \label{fig:model_pipeline}
\end{figure*}

\section{Related Work}
\textbf{Visual Event Understanding} aims to recognize, extract, and structure the actions or activities happening in images or videos. Previous studies simply represent visual events as verbs or subject-verb-object triplets~\cite{chao2015hico,das2013thousand,gupta2015visual,kato2018compositional,li2019transferable,sadhu2021visual}. Recent research studies structural and semantic representations of visual events from image-text and video-text pairs such as M2E2~\cite{li2020cross} and VideoM2E2~\cite{chen2021joint}.  
Importantly, the visual situation recognition task aims to identify not only the activity in an image~\cite{yatskar2016situation,pratt2020grounded} or video~\cite{sadhu2021visual}, but also the entities including persons and objects (\ie, semantic roles) associated with the activity. 
Another benchmark, VisualCOMET~\cite{park2020visualcomet}, proposes to depict person-centric images as a graph of commonsense descriptions, including before-event, intent of people, and next-event. Our work focuses on the video event-relation prediction task and investigates the roles of event type, argument roles, and video features in relation prediction. We also explore using VisualCOMET as an external visual knowledge base for pretraining.

\textbf{Structural Symbolic Representation (SSR)} denotes the representation consisting of discrete tokens with certain structures. SSR has been applied in various tasks in computer vision and natural language processing (NLP). For instance, the visual scene in visual question answering can be modeled as the structural representations of objects with their associated attributes and locations~\cite{yi2018neural}. For event representation, spatio-temporal scene graphs~\cite{ji2020action,rai2021home} decompose each event as a temporal sequence of spatial scene graphs. Recent NLP studies show the potential of SSR in event-relation prediction in text, \eg part-of-speech (POS) and XML tags~\cite{dligach2017neural}. In this work, we follow VidSitu~\cite{sadhu2021visual} to represent each video event as event type/verb and its associated argument roles and entities.

\textbf{Event-relation Prediction in both Text and Video}. Events in texts are often ordered by temporal relation~\cite{chambers2008unsupervised}, causal relation~\cite{mostafazadeh2016caters}, or narrative order~\cite{jans2012skip}. Hong \textit{et al.}~\cite{hong2016building} define event-relations in text as 5 main types, along with 21 sub-types, covering inheritance, expansion, contingency, comparison, and temporality. As the only available video event-relation dataset in cross-document event-relations, VidSitu~\cite{sadhu2021visual} defines four types of relations: no relation, causality, enabled, and reaction-to. Here, we follow the same definition for video event-relation prediction.





\section{Technical Approach}

\subsection{Preliminaries}
\noindent \textbf{Structural Symbolic Representation (SSR).}
SSR generally refers to a representation where elements are discrete tokens and have certain structures, \eg scene graphs~\cite{ji2020action}. To effectively represent an event, the event type (usually a verb), its associated argument roles, and the actual entities for each argument role are required. Therefore, we consider the sequence of text tokens with the following structure as the SSR of an event $x$: $x=\{v, a_1, e_1, a_2, e_2, ..., a_M,e_M\}$, where $v\in \mathbb{W}$ is the event type/verb, $a_m\in \mathbb{W},e_m\in \mathbb{W}$ are the $m^{\text{th}}$ argument role and associated entity, and $M$ is the number of argument roles for this event. Such a sequence is essentially a traverse of the graph with $v$ as the root node and $a_m,e_m$ as the $m^{\text{th}}$ edge and leaf node. 

\noindent \textbf{Event-Relation Prediction.}
Currently VidSitu~\cite{sadhu2021visual} is the only dataset available for video event-relation prediction, and thus we adopt its specific settings. Each video event sequence consists of five consecutive events $\{x_1,x_2,x_3,x_4,x_5\}$, each of which is from a two-second video segment $y_i\in \mathbb{R}^{H \times W \times 3 \times F}$. $H,W,F$ are the height, width, and number of frames of the video segment. In VidSitu, only the relationship between the center event $x_3$ and other events $x_i,i\not=3$ are annotated.

\noindent \textbf{Baseline Model.}
In order to predict the relationship between two events $x_i$ and $x_j$, we follow the RoBERTa variant~\cite{sadhu2021visual} while constructing the baseline model. Based on SSR, a model $\mathcal{F}: \mathbb{W}^{L} \longrightarrow \mathbb{R}^{C}$ takes a sequence of text tokens as input and predicts a distribution over the $C$ possible relationship classes, where $L$ is the length of the sequence consisting of symbolic text representations of $x_i$ and $x_3$. The model $\mathcal{F}$ is initialized with pretrained weights from RoBERTa~\cite{liu2019roberta}.

\noindent \textbf{Baseline + Video Features.}
The state-of-the-art model~\cite{sadhu2021visual} claims that video features are more effective than directly using symbolic representations. It takes both video features and text tokens as inputs $\mathcal{G}: \mathbb{W}^{L} \times \mathbb{R}^{D\times F} \longrightarrow \mathbb{R}^{C}$ to predict the distribution over the $C$ classes. An off-the-shelf video feature extractor $H$ is used to extract continuous feature vectors from the video segment $y_i$ when the event happens. The feature vector is concatenated with the output embedding from the text tokens before being fed into the final classifier $\mathcal{G}$. We denote it as {\sl Baseline + Video Features} in the following discussion. When not specified, the video feature extractor is SlowFast~\cite{feichtenhofer2019slowfast,sadhu2021visual}.

\begin{figure*}[h!]
    \centering
    \includegraphics[width=0.85\textwidth]{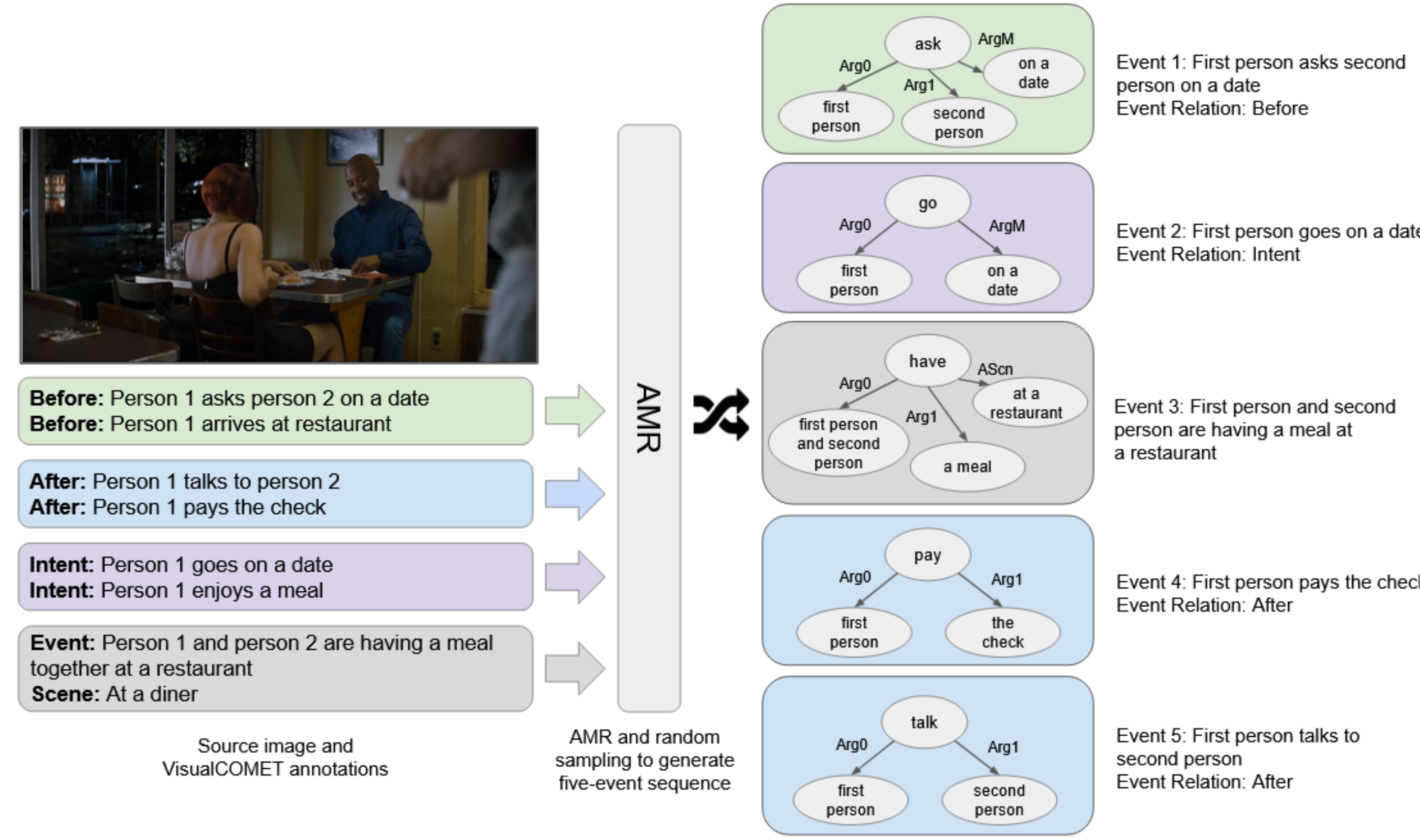}
    \caption{VisualCOMET annotation conversion pipeline. We parse and restructure the full sentence annotations in VisualCOMET using AMRLib~\cite{honnibal2017spacy} into SSRs to match VidSitu annotations for pretraining. Entity co-references are converted from numeric labels to free-form text (e.g. Person 1 $\rightarrow$ First person) to more closely resemble VidSitu.}
    \label{fig:viscom_pipeline}
\end{figure*}

\subsection{Contextualized Event-Sequence Model}
\label{csm}
Rich contextual information (e.g. neighboring events, location of the events, manner of the events, etc.) plays an important role in understanding the relationship between events. We extensively analyze event sequences in the VidSitu dataset and indeed find patterns, as presented in Sec.~\ref{data_analysis}. We propose a novel contextualized Event-Sequence model for event-relation prediction and explore various ways of utilizing context. 

\noindent \textbf{Event-Sequence Model.}
Instead of only feeding the model with two events between which to predict event relations, we propose to exploit the rich contextual information in a full sequence of all the five events (shown in Figure~\ref{fig:model_pipeline}),
\begin{equation}
    p=\mathcal{F}(x_1,x_2,x_3,x_4,x_5),
\end{equation}
where $p\in \mathbb{R}^C$ is the predicted distribution over the $C$ classes of relations. Note that to inform the model between which two events we want to predict the relation, an extra special token ``*'' is added before each of them.

\noindent \textbf{Event-Sequence Model + Video Features.} 
Video features, the continuous feature vectors obtained from pretrained video feature extractors, may convey fine-grained information about the actual visual scene. We follow ~\cite{sadhu2021visual} to integrate video features as in,
\begin{equation}
    p=\mathcal{G}(x_1,x_2,x_3,x_4,x_5,\mathcal{H}(y_i),\mathcal{H}(y_3) ),
\end{equation}
where the video features are fused with contextualized embeddings before the final classification layer.

\noindent \textbf{Sequence-to-Sequence Model.}
Inspired by recent advances of sequence-to-sequence modeling~\cite{T5,lewis2019bart,lin2021vx2text} on both language and video domains, we explore a variant of directly generating the sequence of relationships given the sequence of events as input,
\begin{equation}
    p_{1,3},p_{2,3},p_{4,3},p_{5,3}=\mathcal{S}(x_1,x_2,x_3,x_4,x_5),
\end{equation}
where $p_{i,3}$ is the predicted distribution for the relationship between the $i^{\text{th}}$ event and the middle event $x_3$. The common strategy of teacher-forcing~\cite{williams1989learning} is employed to handle the conditional generation problem, which means during training
\begin{equation}
    p_{k,3}=\mathcal{S}(x_1,x_2,x_3,x_4,x_5, l_1,...,l_{k-1}),
\end{equation}
the ground-truth (GT) ``historical'' event-relations $l_1,...,l_{k-1}$ are used for predicting the next relation. During testing, we apply beam search to decode the actual event relation sentence. 
This variant does not directly use additional contextual information. Instead, it is leveraging conditional generation as a constraint to prevent the model from only predicting the dominant class as reported in ~\cite{sadhu2021visual}.

\noindent \textbf{Auxiliary Arguments.} 
The model in~\cite{sadhu2021visual} only uses base arguments (arguments tied to the verb through direct semantic relations) such the agent and the target. However, additional contextual information could be clearly provided by extra argument roles like AMnr, ADir, and AScn (manner, scene and direction). We append them after the base arguments as additional inputs to the model.


\subsection{Training}
The standard training objective states to use cross-entropy loss to train the model for event-relation prediction,
\begin{equation}
    \min_\theta \text{-} \log p_l,
\end{equation}
where $\theta$ is the parameter to be updated in the model, $l$ is the GT relation, and $p$ is the predicted relation type.

\noindent \textbf{Balancing Data/Loss.} 
In~\cite{sadhu2021visual}, the poor performance of the SSR model is attributed to the possible imbalanced relation class distribution, which leads the model to only predict the dominant relation type. To tackle this issue, we explore two solutions: re-constructing a balanced dataset or using a balanced loss.

For the balanced dataset, we aim at keeping the same number of event pairs in each class by removing videos containing multiple samples of dominant relations. After this process, about 70\% of the dataset is kept.

For the balanced loss, we adopt the commonly used weighted cross entropy loss for optimization,
\begin{equation}
    \min_\theta -\beta_l \log p_l,
\end{equation}
where we set $\beta_l$ as the inverse of the proportion of this relation $l$ in the training set.

\noindent \textbf{VisualCOMET Pretraining.} 
VisualCOMET~\cite{park2020visualcomet} contains 1.4 million commonsense inferences for current visual events under three types of event relations: \textit{Before}, \textit{Intent} and \textit{After}, which correspond to inferring the past, reason, or future event. We reformulate VisualCOMET for event relation prediction pretraining. The main challenge is the different annotation formats between VisualCOMET~\cite{park2020visualcomet} and VidSitu~\cite{sadhu2021visual}. As illustrated in Figure \ref{fig:viscom_pipeline}, VisualCOMET provides natural sentence annotations while VidSitu relies on SSR with verbs and argument roles. We employ AMRLib~\cite{honnibal2017spacy}\footnote{\url{https://github.com/bjascob/amrlib}} to convert VisualCOMET annotations into Abstract Meaning Representation(AMR)~\cite{amr} graphs where the necessary verbs and arguments are restructured to match native VidSitu annotations. Note that the AMR parsing step is crucial in improving performance on VidSitu. We take the current event as $x_3$ and then randomly sample from the three types of annotated events to construct an event sequence $\{x_1,x_2,x_3,x_4,x_5\}$. The relation label is then automatically generated from the type of annotations. Note that during random sampling, to simulate a real event sequence, we restrict $x_1,x_2$ to be either \textit{Before} or \textit{Intent} events and restrict $x_4,x_5$ to be \textit{Intent} or \textit{After} events.


\begin{figure}[!htb]
    \centering
    
    \begin{subfigure}{0.23\textwidth}
        \centering
        \includegraphics[width=1\linewidth]{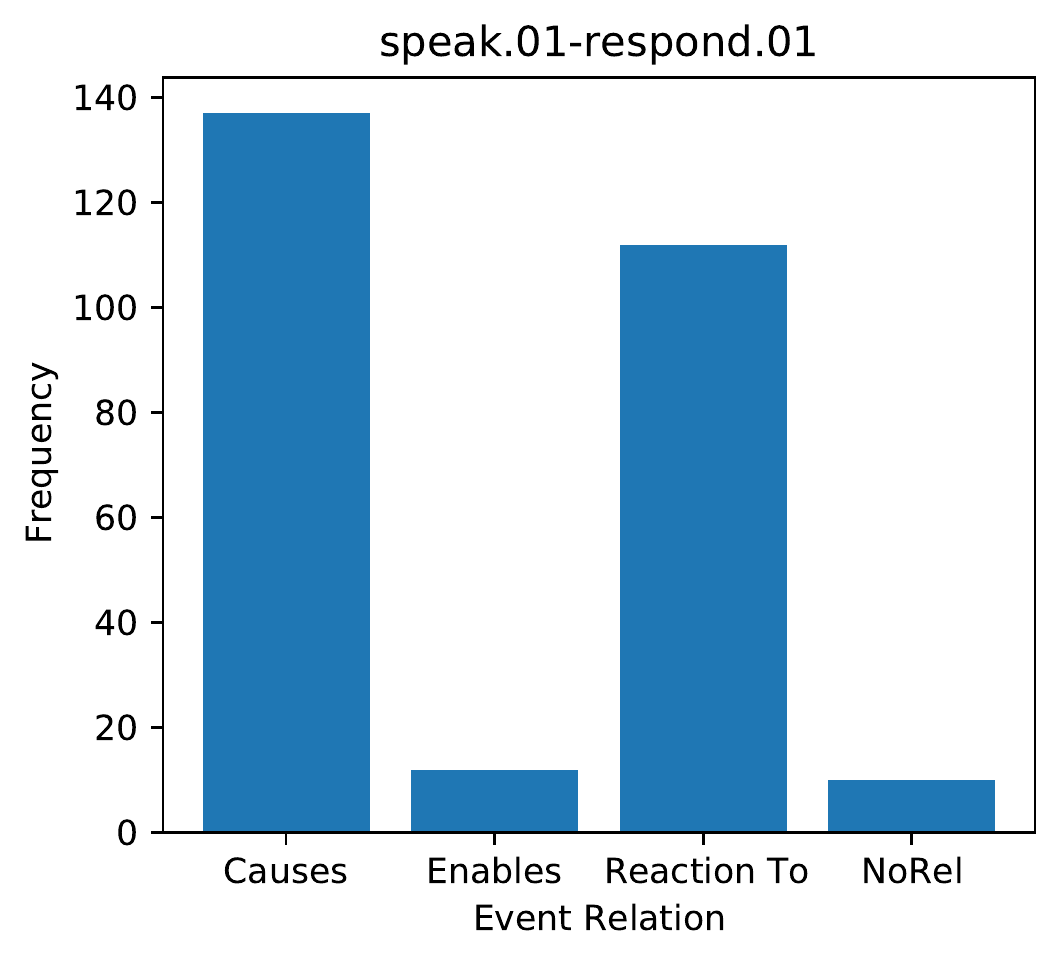}
        \caption{Speak-Respond}
        \label{fig:r1}
    \end{subfigure}
    \begin{subfigure}{0.23\textwidth}
        \centering
        \includegraphics[width=1\linewidth]{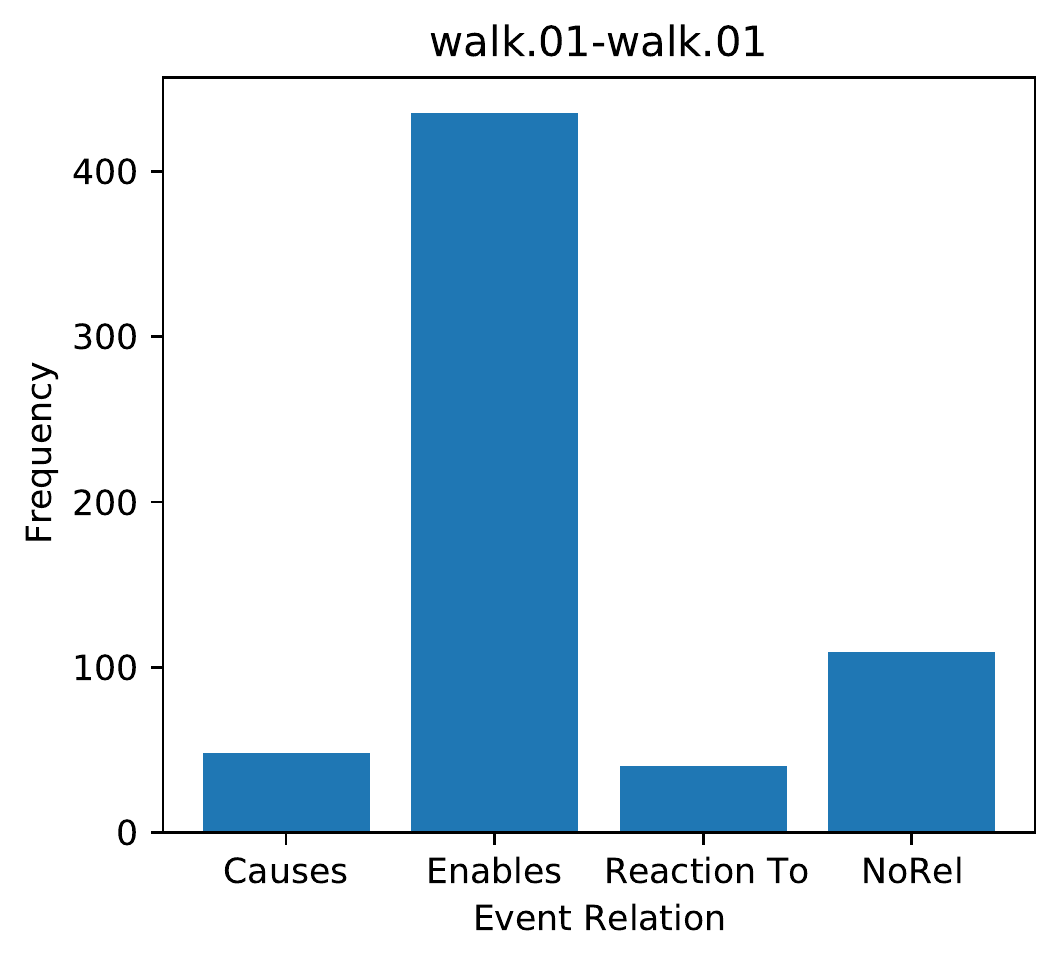}
        \caption{Walk-Walk}
        \label{fig:r2}
    \end{subfigure}
    \caption{ Histograms of the event relations for two event type pairs.}
    \label{fig:event_pair}
\end{figure}

\begin{table*}[th]

\centering
\footnotesize
\begin{tabular}{lcccc}
\toprule
\multicolumn{1}{c}{Model} & LR & Data & Val Macro Top 1 Acc(\%)  & Val Top 1 Acc(\%)
\\ \midrule
 Baseline ~\cite{sadhu2021visual} & 1e-4 & Original & 25.00 & 39.43 
 \\
 Baseline  & 1e-4 &  Balanced & 25.00 & 39.43
 \\
 Baseline + Video Features ~\cite{sadhu2021visual}  & 1e-4 & Original  & 33.73 & 43.71
 \\
 Baseline & 1e-5 & Original & 53.61 & 58.92
 \\
 Baseline  & 1e-5 & Original + balanced loss & 53.98 & 60.47
  \\
 Baseline + Video Features  & 1e-5 & Original & 44.08 & 49.75
 \\
 Event-Sequence & 1e-5 & Original + balanced loss & 55.38 & 61.59
 \\
 
\bottomrule
\end{tabular}
\caption{Comparison between baseline model trained with original unbalanced data, baseline model trained with data balanced on event relation class frequency through undersampling, and baseline model with adjusted learning rate trained on original data.}
\label{balanced_table}
\end{table*}

\section{Experiment Results and Discussions}

\subsection{Dataset and Evaluation Metric}

Our main experiments are conducted on VidSitu~\cite{sadhu2021visual}, a large-scale dataset containing 29,000 10-second video clips where each video clip is divided into five 2-second segments and the most salient verbs and arguments are annotated along with the most dominant event relation. We evaluate event-relation prediction by computing Top-1 accuracy on predicted relations as well as Top-1 accuracy macro-averaged across the four different relation classes: \textit{Causes}, \textit{Enables}, \textit{Reaction To}, and \textit{No Relation}~\cite{sadhu2021visual}.

We adopt VLEP~\cite{lei2020more} to evaluate the generalization of our model on future event prediction, which is formulated as a multiple-choice problem. We follow the official setting and report accuracy on the validation set.

\subsection{Why SSR-Based Methods Fail}
\paragraph{Preliminary Analysis.}
\label{data_analysis}
To understand why SSR baselines~\cite{sadhu2021visual} fail on the VidSitu dataset, we check whether there are event-pairs that have a dominant relation other than \textit{Enables}, which is the overall dominant relation in the dataset. As the two examples in Figure~\ref{fig:event_pair} demonstrate, we found that 66\% of event type pairs satisfy this constraint, which indicates that if the model simply memorizes the dominant event relation for each event type pair, it should achieve a macro-accuracy higher than 25\%~\cite{sadhu2021visual}. This evidence motivates us to further explore different hyperparameters to investigate if the failure comes from the optimization side.

\begin{figure}[!htb]
    \centering
    
    \begin{subfigure}{0.45\textwidth}
        \centering
        \includegraphics[width=1\linewidth]{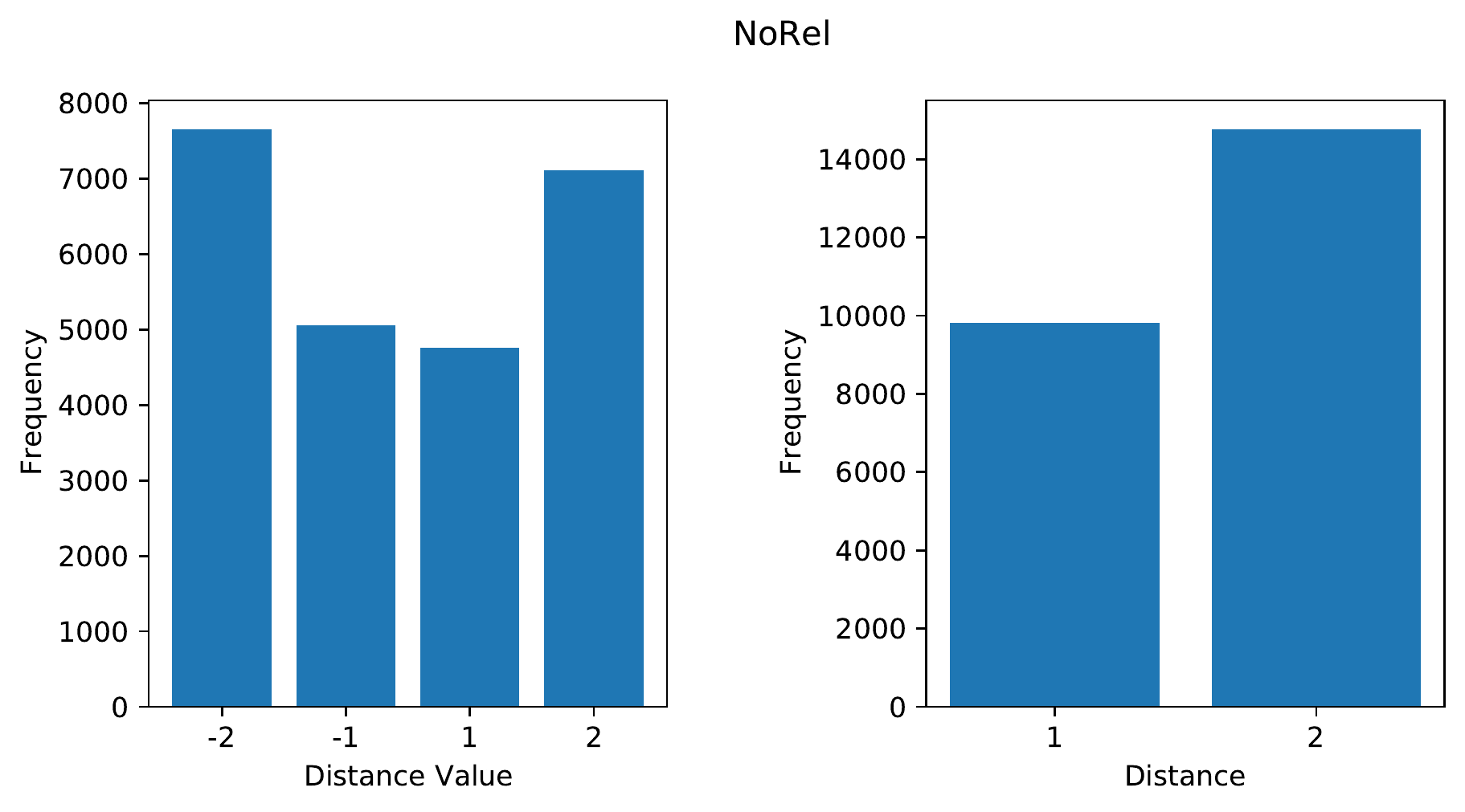}
        \caption{No Relation}
        \label{fig:dist_norel}
    \end{subfigure}
    \begin{subfigure}{0.45\textwidth}
        \centering
        \includegraphics[width=1\linewidth]{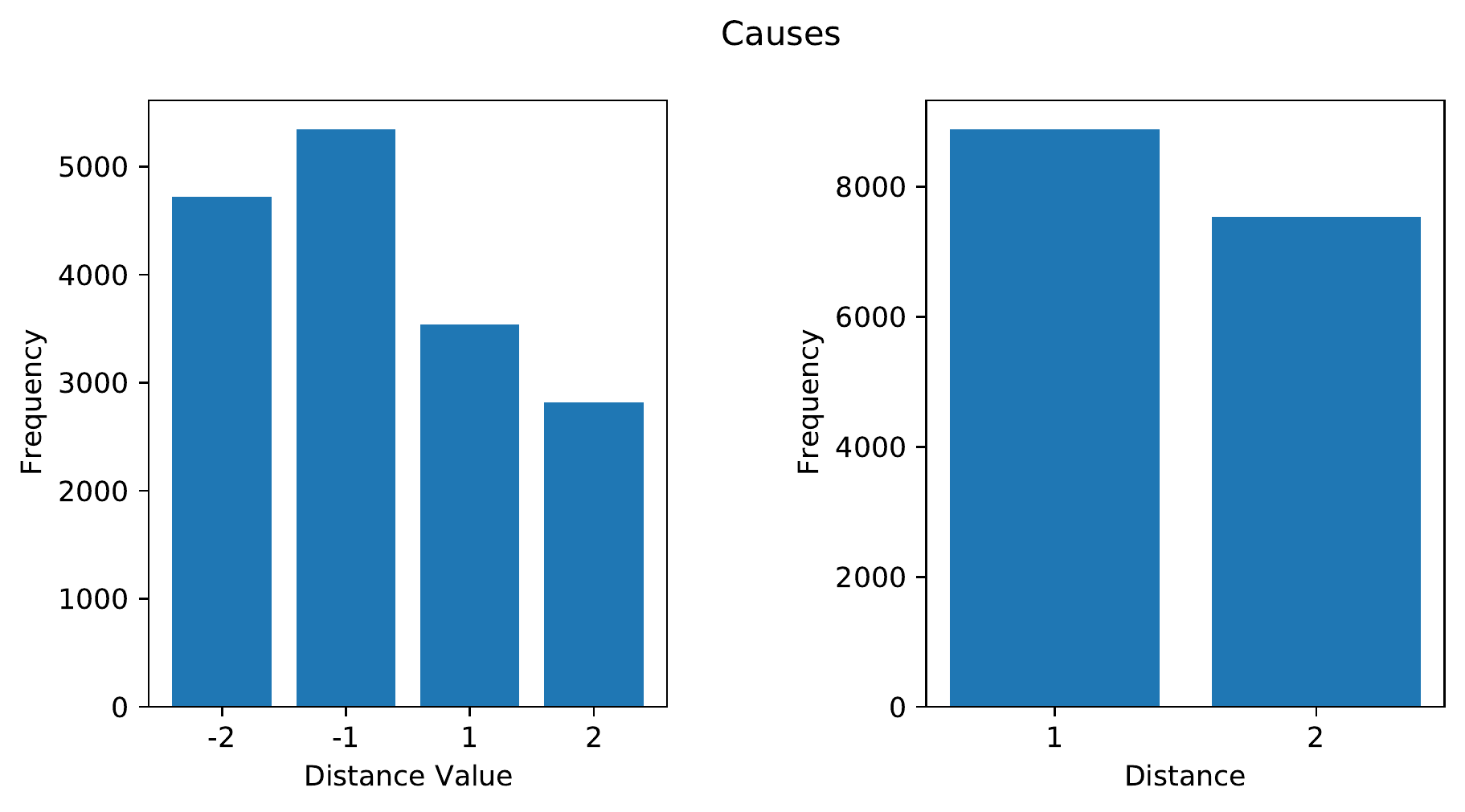}
        \caption{Causes}
        \label{fig:dist_cause}
    \end{subfigure}
    \caption{ Distribution over the relative distance for No Relation and Causes. For example, $x_1$ to $x_3$ has a distance value of -2 and a distance of 2.}
    \label{fig:dist}
\end{figure}

To explore the benefits of using sequences of events as inputs, we study the patterns between the relation type and the relative distance from $x_3$ to the event. As shown in Figure~\ref{fig:dist}, the distance distributions of \textit{No Relation} and  \textit{Causes} are quite different: \textit{No Relation} has an almost symmetric distribution w.r.t. the distance value and is significantly more frequent when the distance is 2; however, \textit{Causes} has a clear peak at distance -1 and decays when the distance value increases, which is consistent with the temporal order. This study verifies that feeding the full sequence of events as inputs is beneficial to the SSR-based models that only predict \textit{Enables}.

\vspace{3pt}
\noindent \textbf{Optimizing Baseline Training Configurations.} This study exploits the potential training configuration issues of the baseline models that reportedly failed to predict meaningful event relations~\cite{sadhu2021visual}.
Since a moderate imbalance is observed in the distribution of event relation classes, we reevaluate the baseline model after training on the same dataset class-balanced through undersampling. This does not improve baseline model performance, which still always predicts one relation class for all events scoring 25\% validation accuracy.

We then discover the reason behind the poor performance on the state-of-the-art baseline models to be a sub-optimally adjusted learning rate rather than lack of patterns in the data. As shown in Table~\ref{balanced_table}, by simply adjusting the learning rate from 1e-4 to 1e-5, performance dramatically improves. Similar adjustments on state-of-the-art multimodal models using text annotations and video features also yield a significant performance improvement. But it is noteworthy that the improvement is much less significant compared to our baseline model, which is built upon SSRs of events.

\subsection{Evaluation Challenges}
\begin{figure*}[h]
    \centering
    \includegraphics[width=0.8\textwidth]{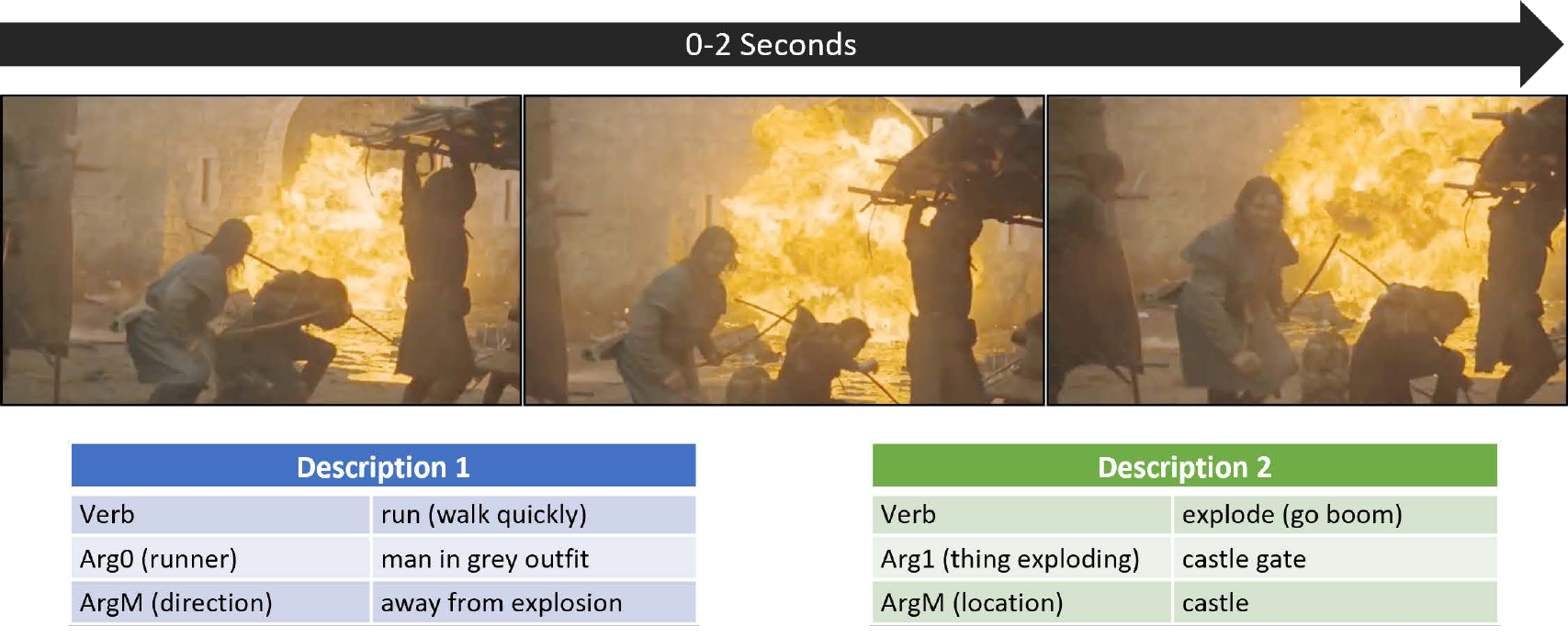}
    \caption{Example scene containing multiple salient events. A large explosion dominates the background while a man is running away from the explosion in the foreground. The verb and argument labels are highly dependent on which event is selected.}
    \label{fig:multi_verb}
\end{figure*}

\paragraph{Simultaneous Events in Videos.}
Despite being rich in context, the noisy nature of video features may actually degrade performance. In complex and busy events, multiple salient verbs may emerge to dominate a scene. In such a scenario, the specific verb chosen drastically influence argument labeling of entity co-references and subsequent event-relationships. Figure~\ref{fig:multi_verb} depicts such a scene where multiple relevant verbs and their corresponding arguments emerge. 
Since event relation annotations are directly tied to annotated event types and argument roles, many instances occur where predicted event relations are accurate to the scene, but evaluated as incorrect since they differ from the annotated events. We explore this further in Section 4.4.

\vspace{3pt}
\noindent \textbf{Quantifying the Effect of Simultaneous Events.}
We evaluate two state-of-the-art video-language models, HERO~\cite{li2020hero} and ClipBERT~\cite{lei2021less}, on video event-relation prediction. We leverage pretrained CLIP-VIT-32~\cite{radford2021learning} to perform frame selection and region selection given event and argument role as the textual input, respectively. In both models, as shown in Table~\ref{clip-selection}, we observe a performance increase when frame selection and region selection are utilized, which indicates a significant presence of irrelevant frames in one video segment and also shows the negative effect cause by simultaneous events.

\label{sec:challenge}

\begin{table}[th]
\centering
\footnotesize
\begin{tabular}{@{}lccc@{}}
\toprule
\multicolumn{1}{c}{Model} & Frame-level & Region-level & Val Macro Acc(\%) 
\\ \midrule
 HERO~\cite{li2020hero} & \xmark & \xmark & 42.15
 \\
 HERO & \cmark & \xmark & 48.03
 \\
 HERO & \cmark & \cmark & 48.48
 \\ \midrule
 ClipBERT~\cite{lei2021less} & \xmark & \xmark & 47.62
 \\
 ClipBERT & \cmark & \xmark & 49.76
 \\
 ClipBERT & \cmark & \cmark & 50.52
 \\
\bottomrule
\end{tabular}
\caption{Performance of HERO and ClipBERT when with and without Frame Selection or Region Selection based on CLIP.}
\label{clip-selection}
\end{table}

\subsection{Improving SSR-Based Methods}

\paragraph{Adding Additional Contextual Information.}

\begin{table*}[h]
\centering

\footnotesize
\begin{tabular}{lcccc}
\toprule

\multicolumn{1}{c}{Model} & Verbs & Base Args & Aux Args & Val Macro Acc (\%)
\\ \midrule
 Baseline (1e-5 lr) & \cmark & \cmark & \xmark & 53.85
 \\
 Baseline + Video Features  (1e-5 lr) & \cmark & \cmark & \xmark & 44.08
 \\
 Event-Sequence Only Verb & \cmark & \xmark & \xmark & 42.53
 \\
 Event-Sequence Only Args & \xmark & \cmark & \cmark & 34.73
 \\
 Event-Sequence Baseline & \cmark & \cmark & \xmark & 55.38
 \\
 Event-Sequence All Args & \cmark & \cmark & \cmark & 58.60
 \\
 Event-Sequence +  vid features (SlowFast) & \cmark & \cmark & \cmark & 55.64
 \\
 Event-Sequence +  vid features (CLIP) & \cmark & \cmark & \cmark & 45.98
 \\ \midrule
 HERO~\cite{li2020hero} & \cmark & \cmark & \cmark & 42.15
 \\
 ClipBERT~\cite{lei2021less} & \cmark & \cmark & \cmark & 47.62
 \\
\bottomrule
\end{tabular}
\caption{Comparison of models given inputs with various forms of context.}
\label{ablation_table}
\end{table*}

When predicting the relationship between two events within a five event sequence from VidSitu, baseline models only take the two target events as inputs. Richer context can be provided directly by increasing the temporal duration of the inputs. As shown in Table~\ref{ablation_table}, by giving the model all five events within a sequence rather than just the two target events, we are able to leverage patterns seen in distances between events as well as events preceding and succeeding the target events, resulting in an accuracy improvement (53.85\% $\rightarrow$ 55.38\%).

Previous state-of-the-art baselines also receive inputs containing only verbs and base arguments (arguments tied to the verb through direct semantic relations). Providing specific context in the form of auxiliary arguments describing the scene of the event, mannerisms and goals of entities performing actions, etc., yields further performance increases (55.38\% $\rightarrow$ 58.60\%). When evaluating event-sequence models given only verbs and only base arguments, performance is still substantially better than random guessing, showing that verbs alone without entity co-references and vice versa provide useful information for event-relation prediction.

Adding context through video features degrades performance rather than showing improvement. We observe a sizable decrease in performance in both the learning rate adjusted baseline (53.58\% $\rightarrow$ 44.08\%) and event-sequence input models (58.60\% $\rightarrow$ 55.64\%) (Table~\ref{ablation_table}). We also observe a decrease in performance when switching to a CLIP-based feature extractor on video inputs instead of SlowFast. In addition, we compare with more powerful vision-language cross-modal baselines ( HERO~\cite{li2020hero} and ClipBERT~\cite{lei2021less}) that have recently shown promise on other video event understanding tasks. However, neither HERO nor ClipBERT perform as well as our Event-Sequence model, which verifies its strong effectiveness.

\begin{figure*}[h!]
    \centering
    \includegraphics[width=0.9\textwidth]{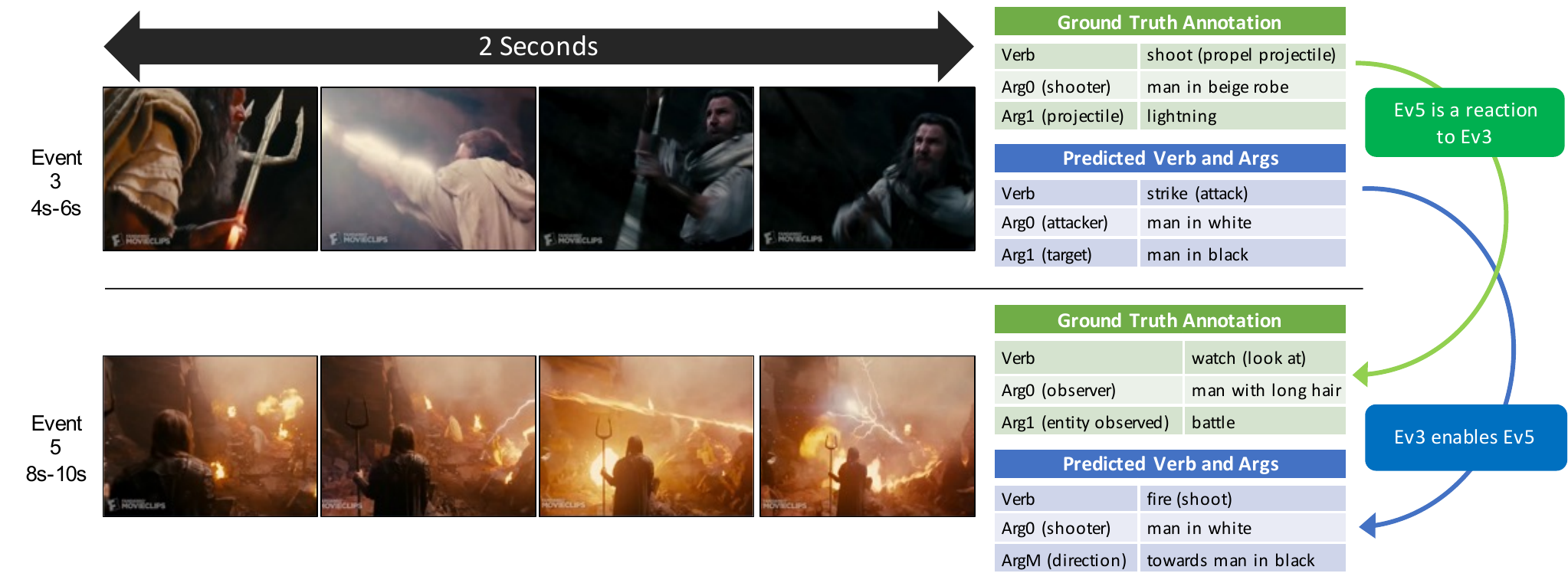}
    \caption{Example where predicted events differ from ground truth human annotations, but predicted verbs and arguments are salient since they describe different events within the same scene. The resulting event-relation prediction using predicted events is also ``accurate'' but different from the ground truth relation since they are relating different, equally salient event descriptions.}
    \label{fig:multi_relations}
\end{figure*}

\begin{table}[th]

\centering
\footnotesize
\begin{tabular}{lcc}
\toprule
\multicolumn{1}{c}{Model} & Val Macro Top 1 Acc & Val Top 1 Acc
\\ \midrule
 Event-Sequence  & 55.38\% & 61.59\%
 \\ \hline
Seq-to-Seq & 52.91\% & 57.43\%
 \\ \hline
 Event-Sequence + All Args & 58.60\% & 62.01\%
 \\ 
  \quad + VisCom pretraining & 59.21\% & 62.65\% 
 \\
\bottomrule
\end{tabular}
\caption{Comparison between Event-Sequence model, Sequence-to-Sequence (Seq-to-Seq) model, and Event-Sequence model pretrained on VisualCOMET data.}
\label{contraints_table}
\end{table}

\vspace{3pt}
\noindent \textbf{Additional Model Constraints.} We find that adding further constraints to event prediction models by employing Sequence-to-Sequence models using BART~\cite{lewis2019bart} does not improve performance beyond our best event-sequence RoBERTa models (Table~\ref{contraints_table}). However, pretraining on VisualCOMET~\cite{park2020visualcomet} yields further improvement, showing that commonsense knowledge of event evolution in VisualCOMET could further help our contextualized sequence model in reasoning event relations.



\begin{table}[th]

\centering
\footnotesize
\begin{tabular}{lcc}
\toprule
\multicolumn{1}{c}{Input} & Val Macro-Acc & Test Macro-Acc
\\ \midrule 
 Annotated verb + args & 55.38\% & 54.47\%
  \\ \hline
Annotated verb + Predicted args & 43.30\% & 42.75\%
\\ \hline
Predicted verb + args & 35.46\% & 34.94\%
 \\ \hline
 Vid features (SlowFast) & 33.78\% & 30.54\%
 \\ 
\bottomrule
\end{tabular}
\caption{Comparison between Event-Sequence models trained and evaluated on human annotations versus predicted events.}
\label{pred_table}
\end{table}

\vspace{3pt}
\noindent \textbf{Performance Analyses Between Predicted and Human Annotations.} To demonstrate an end-to-end approach, we train and evaluate our event-sequence model using predicted verb and argument roles generated by employing a joint feature extractor and text encoder on the video frames~\cite{sadhu2021visual}. Results comparing the models with oracle information, predicted events, and previous state-of-the-art video only baselines~\cite{sadhu2021visual} are summarized in Table~\ref{pred_table}. Notably, our event-sequence model using predicted events outperforms previous video encoder baselines showing the promise of SSRs even when they are generated. We observe a significant improvement when annotated verbs are used to generate argument roles, which verifies the effectiveness of SSR and indicates the ``noisy'' nature of the task and dataset.

Specifically, many scenes occur where predicted verbs and arguments describe different events than ground-truth(GT) human annotations. In the example shown in Figure~\ref{fig:multi_relations}, both GT and predicted events accurately describe the same scene in Event 3: the two men fighting. In Event 5, the predicted verb and arguments describe the background battle and subsequently classifies Event 3 as \textit{Enabling} Event 5 since the battle is a continuation of the previous fighting. However, human annotations describe the man in the foreground watching the battle in the background, and thus the GT label is a \textit{Reaction To} the fighting in Event 3. In this case, \textit{Enables} is an accurate relationship descriptor but considered incorrect. Such discrepancies between annotations likely explain some of the performance degradation when using predicted events.

\subsection{Applications in Downstream Tasks}
We adopt the RoBERTa-based model from \cite{lei2020more} as the baseline, which achieves 65.8\% accuracy on the VLEP validation. It is a challenging task, as prertaining RoBERTa on ATOMIC~\cite{sap2019atomic} only improves accuracy by $0.5\%$. However, when using our event-sequence model as the pretrained weights for the RoBERTa model, we observe a $0.9\%$ improvement of accuracy, which makes video event-relation prediction a better knowledge source of pretraining than the comprehensive textual knowledge base, ATOMIC. This again verifies the effectiveness and the generalization capabilities of our proposed contextualized sequence model.


\section{Conclusion}
In this paper, we defend the effectiveness of SSRs for video event-relation prediction and identify sub-optimal training configurations as the key reason previous models fail. We further provide in-depth analyses and find that video features are noisy because of simultaneous events and irrelevant backgrounds. Oracle event information is important to ensure proper evaluation. We propose a contextualized sequence model with a pretraining technique on VisualCOMET to demonstrate the effectiveness of SSR, which significantly outperforms the state-of-the-art models. Our model trained on video event-relation prediction can be generalized to downstream tasks such as future video event prediction.


\section*{Acknowledgement}
This paper is based upon work supported by U.S. DARPA KAIROS Program No. FA8750-19-2-1004. The views and conclusions contained herein are those of the authors and should not be interpreted as necessarily representing official policies, either expressed or implied, of DARPA, or the U.S. Government. 


{\small
\bibliographystyle{ieee_fullname}
\bibliography{egbib}

\begin{thebibliography}{10}\itemsep=-1pt

\bibitem{ayyubi2022multimodal}
Hammad~A Ayyubi, Christopher Thomas, Lovish Chum, Rahul Lokesh, Yulei Niu,
  Xudong Lin, Long Chen, Jaywon Koo, Sounak Ray, and Shih-Fu Chang.
\newblock Multimodal event graphs: Towards event centric understanding of
  multimodal world.
\newblock {\em arXiv preprint arXiv:2206.07207}, 2022.

\bibitem{badgett2016extracting}
Allison Badgett and Ruihong Huang.
\newblock Extracting subevents via an effective two-phase approach.
\newblock In {\em Proceedings of the 2016 Conference on Empirical Methods in
  Natural Language Processing}, pages 906--911, 2016.

\bibitem{amr}
Laura Banarescu, Claire Bonial, Shu Cai, Madalina Georgescu, Kira Griffitt, Ulf
  Hermjakob, Kevin Knight, Philipp Koehn, Martha Palmer, and Nathan Schneider.
\newblock Abstract meaning representation for sembanking.
\newblock In {\em Proceedings of the 7th linguistic annotation workshop and
  interoperability with discourse}, pages 178--186, 2013.

\bibitem{bosselut2019comet}
Antoine Bosselut, Hannah Rashkin, Maarten Sap, Chaitanya Malaviya, Asli
  Celikyilmaz, and Yejin Choi.
\newblock Comet: Commonsense transformers for knowledge graph construction.
\newblock In {\em Association for Computational Linguistics (ACL)}, 2019.

\bibitem{chambers2008unsupervised}
Nathanael Chambers and Dan Jurafsky.
\newblock Unsupervised learning of narrative event chains.
\newblock In {\em Proceedings of ACL-08: HLT}, pages 789--797, 2008.

\bibitem{chao2015hico}
Yu-Wei Chao, Zhan Wang, Yugeng He, Jiaxuan Wang, and Jia Deng.
\newblock Hico: A benchmark for recognizing human-object interactions in
  images.
\newblock In {\em Proceedings of the IEEE International Conference on Computer
  Vision}, pages 1017--1025, 2015.

\bibitem{chen2021joint}
Brian Chen, Xudong Lin, Christopher Thomas, Manling Li, Shoya Yoshida, Lovish
  Chum, Heng Ji, and Shih-Fu Chang.
\newblock Joint multimedia event extraction from video and article.
\newblock In {\em Findings of the Association for Computational Linguistics:
  EMNLP 2021}, pages 74--88, 2021.

\bibitem{das2013thousand}
Pradipto Das, Chenliang Xu, Richard~F Doell, and Jason~J Corso.
\newblock A thousand frames in just a few words: Lingual description of videos
  through latent topics and sparse object stitching.
\newblock In {\em Proceedings of the IEEE conference on computer vision and
  pattern recognition}, pages 2634--2641, 2013.

\bibitem{dligach2017neural}
Dmitriy Dligach, Timothy Miller, Chen Lin, Steven Bethard, and Guergana Savova.
\newblock Neural temporal relation extraction.
\newblock In {\em Proceedings of the 15th Conference of the European Chapter of
  the Association for Computational Linguistics: Volume 2, Short Papers}, pages
  746--751, 2017.

\bibitem{feichtenhofer2019slowfast}
Christoph Feichtenhofer, Haoqi Fan, Jitendra Malik, and Kaiming He.
\newblock Slowfast networks for video recognition.
\newblock In {\em Proceedings of the IEEE/CVF international conference on
  computer vision}, pages 6202--6211, 2019.

\bibitem{glavavs2014hieve}
Goran Glava{\v{s}}, Jan {\v{S}}najder, Parisa Kordjamshidi, and Marie-Francine
  Moens.
\newblock Hieve: A corpus for extracting event hierarchies from news stories.
\newblock In {\em Proceedings of 9th language resources and evaluation
  conference}, pages 3678--3683. ELRA, 2014.

\bibitem{gupta2015visual}
Saurabh Gupta and Jitendra Malik.
\newblock Visual semantic role labeling.
\newblock {\em arXiv preprint arXiv:1505.04474}, 2015.

\bibitem{hong2016building}
Yu Hong, Tongtao Zhang, Tim O’Gorman, Sharone Horowit-Hendler, Heng Ji, and
  Martha Palmer.
\newblock Building a cross-document event-event relation corpus.
\newblock In {\em Proceedings of the 10th Linguistic Annotation Workshop held
  in conjunction with ACL 2016 (LAW-X 2016)}, pages 1--6, 2016.

\bibitem{honnibal2017spacy}
Matthew Honnibal and Ines Montani.
\newblock spacy 2: Natural language understanding with bloom embeddings,
  convolutional neural networks and incremental parsing.
\newblock {\em To appear}, 7(1):411--420, 2017.

\bibitem{hudson2019learning}
Drew Hudson and Christopher~D Manning.
\newblock Learning by abstraction: The neural state machine.
\newblock {\em Advances in Neural Information Processing Systems}, 32, 2019.

\bibitem{jans2012skip}
Bram Jans, Steven Bethard, Ivan Vulic, and Marie-Francine Moens.
\newblock Skip n-grams and ranking functions for predicting script events.
\newblock In {\em Proceedings of the 13th Conference of the European Chapter of
  the Association for Computational Linguistics (EACL 2012)}, pages 336--344.
  ACL; East Stroudsburg, PA, 2012.

\bibitem{ji2020action}
Jingwei Ji, Ranjay Krishna, Li Fei-Fei, and Juan~Carlos Niebles.
\newblock Action genome: Actions as compositions of spatio-temporal scene
  graphs.
\newblock In {\em Proceedings of the IEEE/CVF Conference on Computer Vision and
  Pattern Recognition}, pages 10236--10247, 2020.

\bibitem{kato2018compositional}
Keizo Kato, Yin Li, and Abhinav Gupta.
\newblock Compositional learning for human object interaction.
\newblock In {\em Proceedings of the European Conference on Computer Vision
  (ECCV)}, pages 234--251, 2018.

\bibitem{lei2021less}
Jie Lei, Linjie Li, Luowei Zhou, Zhe Gan, Tamara~L. Berg, Mohit Bansal, and
  Jingjing Liu.
\newblock Less is more: Clipbert for video-and-language learningvia sparse
  sampling.
\newblock In {\em CVPR}, 2021.

\bibitem{lei2020more}
Jie Lei, Licheng Yu, Tamara~L Berg, and Mohit Bansal.
\newblock What is more likely to happen next? video-and-language future event
  prediction.
\newblock {\em arXiv preprint arXiv:2010.07999}, 2020.

\bibitem{lewis2019bart}
Mike Lewis, Yinhan Liu, Naman Goyal, Marjan Ghazvininejad, Abdelrahman Mohamed,
  Omer Levy, Ves Stoyanov, and Luke Zettlemoyer.
\newblock Bart: Denoising sequence-to-sequence pre-training for natural
  language generation, translation, and comprehension.
\newblock {\em arXiv preprint arXiv:1910.13461}, 2019.

\bibitem{li2020hero}
Linjie Li, Yen-Chun Chen, Yu Cheng, Zhe Gan, Licheng Yu, and Jingjing Liu.
\newblock Hero: Hierarchical encoder for video+ language omni-representation
  pre-training.
\newblock {\em arXiv preprint arXiv:2005.00200}, 2020.

\bibitem{li2020cross}
Manling Li, Alireza Zareian, Qi Zeng, Spencer Whitehead, Di Lu, Heng Ji, and
  Shih-Fu Chang.
\newblock Cross-media structured common space for multimedia event extraction.
\newblock In {\em Proceedings of the 58th Annual Meeting of the Association for
  Computational Linguistics}, pages 2557--2568, 2020.

\bibitem{li2019transferable}
Yong-Lu Li, Siyuan Zhou, Xijie Huang, Liang Xu, Ze Ma, Hao-Shu Fang, Yanfeng
  Wang, and Cewu Lu.
\newblock Transferable interactiveness knowledge for human-object interaction
  detection.
\newblock In {\em Proceedings of the IEEE/CVF Conference on Computer Vision and
  Pattern Recognition}, pages 3585--3594, 2019.

\bibitem{lin2021vx2text}
Xudong Lin, Gedas Bertasius, Jue Wang, Shih-Fu Chang, Devi Parikh, and Lorenzo
  Torresani.
\newblock Vx2text: End-to-end learning of video-based text generation from
  multimodal inputs.
\newblock In {\em Proceedings of the IEEE/CVF Conference on Computer Vision and
  Pattern Recognition}, pages 7005--7015, 2021.

\bibitem{lin2022towards}
Xudong Lin, Simran Tiwari, Shiyuan Huang, Manling Li, Mike~Zheng Shou, Heng Ji,
  and Shih-Fu Chang.
\newblock Towards fast adaptation of pretrained contrastive models for
  multi-channel video-language retrieval.
\newblock {\em arXiv preprint arXiv:2206.02082}, 2022.

\bibitem{liu2020event}
Jian Liu, Yubo Chen, Kang Liu, Wei Bi, and Xiaojiang Liu.
\newblock Event extraction as machine reading comprehension.
\newblock In {\em Proceedings of the 2020 Conference on Empirical Methods in
  Natural Language Processing (EMNLP)}, pages 1641--1651, Online, 2020.
  Association for Computational Linguistics.

\bibitem{liu-etal-2019-neural-cross}
Jian Liu, Yubo Chen, Kang Liu, and Jun Zhao.
\newblock Neural cross-lingual event detection with minimal parallel resources.
\newblock In {\em Proceedings of the 2019 Conference on Empirical Methods in
  Natural Language Processing and the 9th International Joint Conference on
  Natural Language Processing (EMNLP-IJCNLP)}, pages 738--748, Hong Kong,
  China, 2019. Association for Computational Linguistics.

\bibitem{liu2019roberta}
Yinhan Liu, Myle Ott, Naman Goyal, Jingfei Du, Mandar Joshi, Danqi Chen, Omer
  Levy, Mike Lewis, Luke Zettlemoyer, and Veselin Stoyanov.
\newblock Roberta: A robustly optimized bert pretraining approach.
\newblock {\em arXiv preprint arXiv:1907.11692}, 2019.

\bibitem{mostafazadeh2016caters}
Nasrin Mostafazadeh, Alyson Grealish, Nathanael Chambers, James Allen, and Lucy
  Vanderwende.
\newblock Caters: Causal and temporal relation scheme for semantic annotation
  of event structures.
\newblock In {\em Proceedings of the Fourth Workshop on Events}, pages 51--61,
  2016.

\bibitem{nguyen-etal-2016-joint}
Thien~Huu Nguyen, Kyunghyun Cho, and Ralph Grishman.
\newblock Joint event extraction via recurrent neural networks.
\newblock In {\em Proceedings of the 2016 Conference of the North {A}merican
  Chapter of the Association for Computational Linguistics: Human Language
  Technologies}, pages 300--309, San Diego, California, 2016. Association for
  Computational Linguistics.

\bibitem{o2016richer}
Tim O’Gorman, Kristin Wright-Bettner, and Martha Palmer.
\newblock Richer event description: Integrating event coreference with
  temporal, causal and bridging annotation.
\newblock In {\em Proceedings of the 2nd Workshop on Computing News Storylines
  (CNS 2016)}, pages 47--56, 2016.

\bibitem{park2020visualcomet}
Jae~Sung Park, Chandra Bhagavatula, Roozbeh Mottaghi, Ali Farhadi, and Yejin
  Choi.
\newblock Visualcomet: Reasoning about the dynamic context of a still image.
\newblock In {\em European Conference on Computer Vision}, pages 508--524.
  Springer, Cham, 2020.

\bibitem{pratt2020grounded}
Sarah Pratt, Mark Yatskar, Luca Weihs, Ali Farhadi, and Aniruddha Kembhavi.
\newblock Grounded situation recognition.
\newblock In {\em European Conference on Computer Vision}, pages 314--332.
  Springer, 2020.

\bibitem{radford2021learning}
Alec Radford, Jong~Wook Kim, Chris Hallacy, Aditya Ramesh, Gabriel Goh,
  Sandhini Agarwal, Girish Sastry, Amanda Askell, Pamela Mishkin, Jack Clark,
  et~al.
\newblock Learning transferable visual models from natural language
  supervision.
\newblock In {\em International Conference on Machine Learning}, pages
  8748--8763. PMLR, 2021.

\bibitem{T5}
Colin Raffel, Noam Shazeer, Adam Roberts, Katherine Lee, Sharan Narang, Michael
  Matena, Yanqi Zhou, Wei Li, and Peter~J Liu.
\newblock Exploring the limits of transfer learning with a unified text-to-text
  transformer.
\newblock {\em arXiv preprint arXiv:1910.10683}, 2019.

\bibitem{rai2021home}
Nishant Rai, Haofeng Chen, Jingwei Ji, Rishi Desai, Kazuki Kozuka, Shun
  Ishizaka, Ehsan Adeli, and Juan~Carlos Niebles.
\newblock Home action genome: Cooperative compositional action understanding.
\newblock In {\em Proceedings of the IEEE/CVF Conference on Computer Vision and
  Pattern Recognition}, pages 11184--11193, 2021.

\bibitem{sadhu2021visual}
Arka Sadhu, Tanmay Gupta, Mark Yatskar, Ram Nevatia, and Aniruddha Kembhavi.
\newblock Visual semantic role labeling for video understanding.
\newblock In {\em Proceedings of the IEEE/CVF Conference on Computer Vision and
  Pattern Recognition}, pages 5589--5600, 2021.

\bibitem{sap2019atomic}
Maarten Sap, Ronan Le~Bras, Emily Allaway, Chandra Bhagavatula, Nicholas
  Lourie, Hannah Rashkin, Brendan Roof, Noah~A Smith, and Yejin Choi.
\newblock Atomic: An atlas of machine commonsense for if-then reasoning.
\newblock In {\em Proceedings of the AAAI Conference on Artificial
  Intelligence}, volume~33, pages 3027--3035, 2019.

\bibitem{sha2018jointly}
Lei Sha, Feng Qian, Baobao Chang, and Zhifang Sui.
\newblock Jointly extracting event triggers and arguments by dependency-bridge
  {RNN} and tensor-based argument interaction.
\newblock In Sheila~A. McIlraith and Kilian~Q. Weinberger, editors, {\em
  Proceedings of the Thirty-Second {AAAI} Conference on Artificial
  Intelligence, (AAAI-18), the 30th innovative Applications of Artificial
  Intelligence (IAAI-18), and the 8th {AAAI} Symposium on Educational Advances
  in Artificial Intelligence (EAAI-18), New Orleans, Louisiana, USA, February
  2-7, 2018}, pages 5916--5923. {AAAI} Press, 2018.

\bibitem{wang2020neural}
Xiaozhi Wang, Shengyu Jia, Xu Han, Zhiyuan Liu, Juanzi Li, Peng Li, and Jie
  Zhou.
\newblock {N}eural {G}ibbs {S}ampling for {J}oint {E}vent {A}rgument
  {E}xtraction.
\newblock In {\em Proceedings of the 1st Conference of the Asia-Pacific Chapter
  of the Association for Computational Linguistics and the 10th International
  Joint Conference on Natural Language Processing}, pages 169--180, Suzhou,
  China, 2020. Association for Computational Linguistics.

\bibitem{wang2019hmeae}
Xiaozhi Wang, Ziqi Wang, Xu Han, Zhiyuan Liu, Juanzi Li, Peng Li, Maosong Sun,
  Jie Zhou, and Xiang Ren.
\newblock {HMEAE}: Hierarchical modular event argument extraction.
\newblock In {\em Proceedings of the 2019 Conference on Empirical Methods in
  Natural Language Processing and the 9th International Joint Conference on
  Natural Language Processing (EMNLP-IJCNLP)}, pages 5777--5783, Hong Kong,
  China, 2019. Association for Computational Linguistics.

\bibitem{wanglanguage}
Zhenhailong Wang, Manling Li, Ruochen Xu, Luowei Zhou, Jie Lei, Xudong Lin,
  Shuohang Wang, Ziyi Yang, Chenguang Zhu, Derek Hoiem, et~al.
\newblock Language models with image descriptors are strong few-shot
  video-language learners.
\newblock In {\em Advances in Neural Information Processing Systems}.

\bibitem{williams1989learning}
Ronald~J Williams and David Zipser.
\newblock A learning algorithm for continually running fully recurrent neural
  networks.
\newblock {\em Neural computation}, 1(2):270--280, 1989.

\bibitem{xu2019scene}
Ning Xu, An-An Liu, Jing Liu, Weizhi Nie, and Yuting Su.
\newblock Scene graph captioner: Image captioning based on structural visual
  representation.
\newblock {\em Journal of Visual Communication and Image Representation},
  58:477--485, 2019.

\bibitem{yang2022video}
Guang Yang, Manling Li, Xudong Lin, Jiajie Zhang, Shih-Fu Chang, and Heng Ji.
\newblock Video event extraction via tracking visual states of arguments.
\newblock {\em arXiv preprint arXiv:2211.01781}, 2022.

\bibitem{yao2020weakly}
Wenlin Yao, Zeyu Dai, Maitreyi Ramaswamy, Bonan Min, and Ruihong Huang.
\newblock Weakly supervised subevent knowledge acquisition.
\newblock In {\em Proceedings of the 2020 Conference on Empirical Methods in
  Natural Language Processing (EMNLP)}, 2020.

\bibitem{yatskar2016situation}
Mark Yatskar, Luke Zettlemoyer, and Ali Farhadi.
\newblock Situation recognition: Visual semantic role labeling for image
  understanding.
\newblock In {\em Proceedings of the IEEE conference on computer vision and
  pattern recognition}, pages 5534--5542, 2016.

\bibitem{ye2015eventnet}
Guangnan Ye, Yitong Li, Hongliang Xu, Dong Liu, and Shih-Fu Chang.
\newblock Eventnet: A large scale structured concept library for complex event
  detection in video.
\newblock In {\em Proceedings of the 23rd ACM international conference on
  Multimedia}, pages 471--480, 2015.

\bibitem{yi2018neural}
Kexin Yi, Jiajun Wu, Chuang Gan, Antonio Torralba, Pushmeet Kohli, and Josh
  Tenenbaum.
\newblock Neural-symbolic vqa: Disentangling reasoning from vision and language
  understanding.
\newblock {\em Advances in neural information processing systems}, 31, 2018.

\end{thebibliography}
}

\end{document}